\documentclass{CVM}
\usepackage{amsmath,amsfonts}
\usepackage{algorithmic}
\usepackage{algorithm2e}
\usepackage{array}
\usepackage{textcomp}
\usepackage{stfloats}
\usepackage{url}
\usepackage{verbatim}
\usepackage{graphicx}
\usepackage{booktabs}
\usepackage{marvosym}
\usepackage{caption}
\usepackage{lipsum}

\CVMsetup{
type      = {Research Article},
doi       = {s41095-0xx-xxxx-x},
title     = {Anglehedral Surface: Self-supervised Point Cloud Reconstruction based on Anglehedral Surface},
author   = {Hui Tian$^{1}$\cor{}, Kai Xu$^{2}$},
runauthor = {Hui Tian, Kai Xu},
abstract  = {Surface reconstruction from raw point clouds has been a pivotal focus in computer graphics for numerous years, primarily due to its integral role in modeling and rendering applications. A key approach to tackling this challenge involves establishing local geometries to align with the local curve. However, conventional methodologies typically resort to constructing either a local plane or a polynomial curve. While a local plane may result in the loss of sharp features and boundary errors on open surfaces, utilizing a polynomial curve can pose difficulties when integrating with neural networks due to issues with local coordinate consistency. In response to these limitations, we introduce a groundbreaking concept known as anglehedral surfaces to represent local geometry. This innovative method offers enhanced flexibility in portraying sharp features and surface boundaries on open surfaces without necessitating a local coordinate system, which is crucial for neural network integration. Specifically, we leverage normals to construct anglehedral surfaces, encompassing dihedral and trihedral surfaces controlled by 2 and 3 normals, respectively. As a result, our approach adeptly accommodates sharp features on turning surfaces and accurately captures boundaries on open surfaces. To demonstrate superior performance, our method excels on three widely-used datasets (ShapeNetCore, ABC, and ScanNet). Code is available https://gitee.com/HuiTian1/3fold.
},
keywords  = {Anglehedral surface, Point Cloud, Surface Reconstruction, Implicit Surface},
copyright = {The Author(s)},
}

\begin{document}

\maketitle

\begin{figure}[b] \vskip -4mm
    \small\renewcommand\arraystretch{1.3}
        \begin{tabular}{p{80.5mm}} \toprule\\ \end{tabular}
        \vskip -4.5mm \noindent \setlength{\tabcolsep}{1pt}
        \begin{tabular}{p{3.5mm}p{80mm}}
    $1\quad $ & National University of Defense Technology, ChangSha, 410073, China. E-mail: tianhui13@nudt.edu.cn\\
    $2\quad $ & National University of Defense Technology, ChangSha, 410073, China. E-mail: kevin.kai.xu@gmail.com\\
    
&\hspace{-5mm} Manuscript received: 2022-01-01; accepted: 2022-01-01\vspace{-2mm}
    \end{tabular} \vspace {-3mm}
    \end{figure}


\section{Introduction}

\label{intro}
Surface reconstruction from point cloud is a crucial task in computer graphics and computer vision, given its broad applications in modeling and rendering. Recent methods \cite{2020Convolutional, 2022POCO} aim to directly predict the Signed Distance Function (SDF) from the point cloud to achieve an implicit representation, showing promising performance. However, the use of SDF-based techniques is limited by their ability to represent only watertight surfaces. Unsigned Distance Function (UDF) methods can overcome this limitation and extend the representation to open surfaces. Recent studies \cite{2020ndf, superudf, deepmls} have focused on this advancement. To accurately predict the UDF for a query point, constructing differentiable local geometry to fit the surface is essential. Previous works\cite{deepmls, superudf, mls} have predominantly used either plane or polynomial surfaces for local geometry. However, planes often struggle with sharp feature loss and boundary errors, while polynomial surfaces face issues with coordinate consistency. Introducing the concept of "Anglehedral surfaces" not only mitigates errors related to sharp features and boundaries but also addresses the challenge of coordinate consistency.

\begin{figure*} 
\centering
  \includegraphics[width=2\columnwidth]{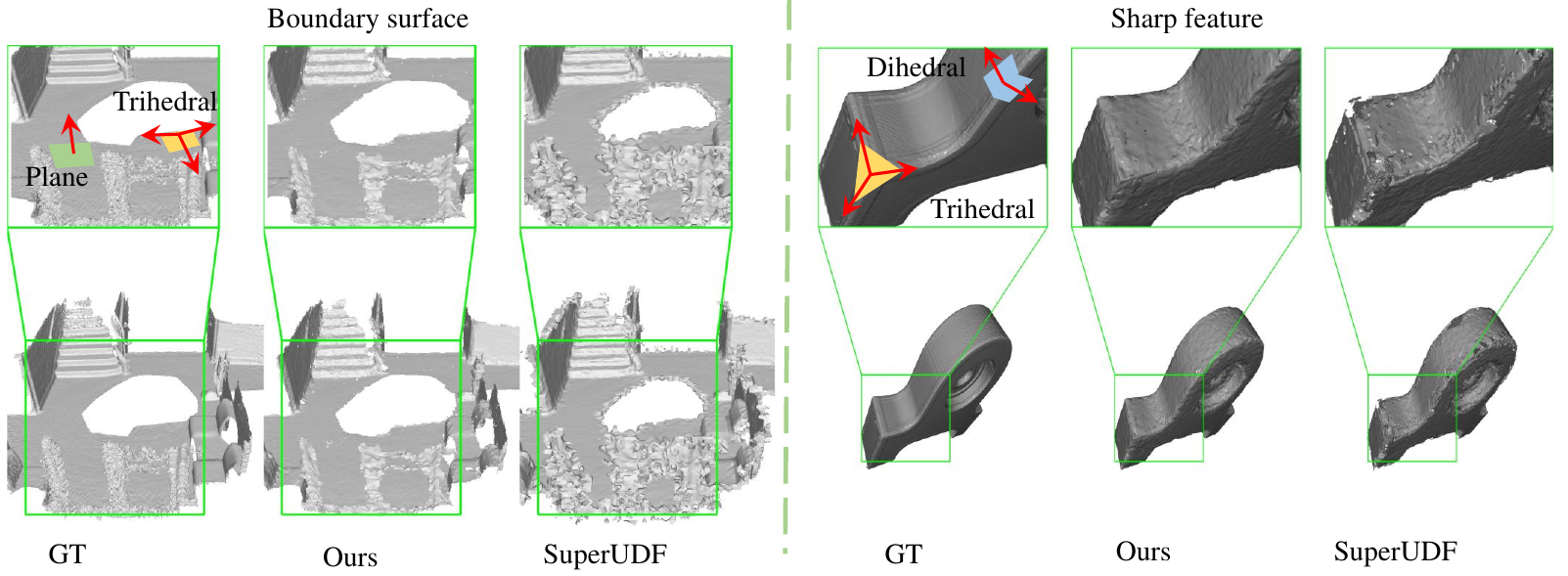}
  \caption{Anglehedral surface often appears in scenes or shapes. The examples separately show the priority of the anglehedral surface in representing the surface boundary when 3 normals lie on the same plane and sharp surface. SuperUDF only makes use of the local geometry plane while ours makes use of the local geometry Anglehedral surface.}
  \label{fig:teaser}
\end{figure*}

\begin{figure*} 
\centering
  \includegraphics[width=2\columnwidth]{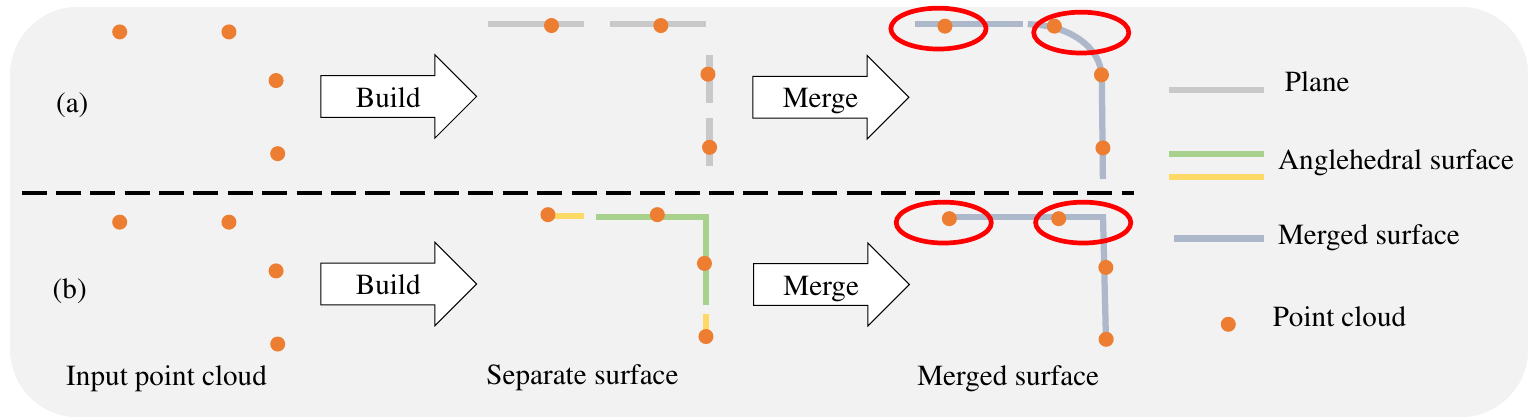}
  \caption{The advantage of anglehedral surface illustrated in 2D cases. (a) is situation where there is only plane. (b) is the situation where there are plane and anglehedral surface. From the mesh of red circle, the mesh represented by anglehedral surface is close to the GT.}
  \label{fig:advantage}
\end{figure*}

First, let's examine the errors present on sharp features and boundary surfaces. In previous research\cite{imls, deepmls, superudf}, a common approach involved calculating a local plane for each point in the point cloud and then merging these planes using distance-based weighting. While the IMLS method\cite{imls} has demonstrated that the representation capability of merging local planes is equivalent to that of polynomial surfaces, it struggles to accurately represent sharp features on curved surfaces. Furthermore, as planes can extend in all directions, they fail to properly capture the edges on the boundary of an open surface. These two issues are illustrated in Fig. \ref{fig:advantage} (a), where the red circle highlights the identified problems. 

Secondly, to construct a local polynomial surface, the initial step involves establishing the local coordinate system. In conventional approaches\cite{imls}, techniques such as PCA (Principal Component Analysis) or similar methodologies are employed to analyze points within a proximity for the creation of the local coordinate system. Subsequently, leveraging this local coordinate system, the parameters of the polynomial surface can be computed. However, when merging this process with neural network integration, direct prediction of the polynomial parameters becomes challenging due to the reliance on the local coordinate system. The neural network is unable to discern how the local coordinate system is constructed, impeding the direct prediction of polynomial parameters. While a potential workaround involves first constructing the coordinate system and then inputting both the coordinate system and the neighborhood point cloud into the neural network, effectively learning the polynomial parameters remains a formidable task.

In order to more effectively depict the boundaries of open surfaces and intricate local geometry without the need for establishing a local coordinate system, we introduce anglehedral surfaces. This innovative approach enables the detailed representation of complex features while circumventing issues related to coordinate consistency. In various shapes and scenes, open surface boundaries and sharp features like turning surfaces or corners are common occurrences, as illustrated in Fig~\ref{fig:teaser}, turning surface means surface where its normal turns, such as the edge of a cube. Anglehedral surfaces offer a natural solution for representing these features. We introduce two types of anglehedral surfaces: dihedral surfaces and trihedral surfaces, as depicted in Fig~\ref{fig:teaser}.

Dihedral surfaces are suitable for representing turning surfaces, utilizing two normals originating from the same point to capture the local geometry, the intersection of two planes defines a unique line that is perpendicular to the two normals, with each normal defining a half plane that together compose the dihedral surface. The variation of the two normals can control the adaptation of the dihedral surface.

For the representation of corners on a cube or open surface boundaries, we employ trihedral surfaces. In this case, three normals are utilized to describe the local geometry, with the trihedral surface adjusting accordingly as the normals vary.

In summary, the utilization of anglehedral surfaces offers two distinct advantages, albeit with a slightly increased computational complexity. Firstly, these surfaces present a more nuanced local descriptor in comparison to traditional planes, enabling a more precise representation of local geometry such as turning surfaces, corner surfaces, and boundary surfaces. Secondly, by eliminating the necessity for a consistent local coordinate system, we can harness the power of neural networks to predict local geometry parameters effectively. What's more, the learning process to optimize the anglehedral surface parameter is differentiable. They can not only represent the surface with sharp feature, but also the surface like cylindrical surface because all the local geometry will be merged.

To demonstrate the efficacy of our approach, we utilize chamfer distance to assess the quality of our reconstructed surfaces. Our method outperforms state-of-the-art methods across all three datasets: ShapeNet, ABC, and ScanNet. Particularly noteworthy is the significant improvement achieved on the ABC dataset, which features shapes with numerous sharp features. Our method demonstrates a remarkable 19\% reduction in chamfer distance on this dataset. We make the following contributions to the field of point cloud reconstruction:
\begin{itemize}
    \item Firstly, we present a innovative method for depicting local surfaces through the introduction of "anglehedral surfaces", encompassing dihedral and trihedral surface representations. These representations adaptively encapsulate the intricate details of local surfaces, delivering a more precise and thorough representation without necessitating a local coordinate system.
    \item Secondly, our approach attains results on par with state-of-the-art techniques in surface reconstruction from point clouds. By harnessing the benefits of anglehedral surfaces, we can generate high-quality reconstructions that rival the top-performing methods in the field.
\end{itemize}

These contributions represent a significant advancement in the field of point cloud reconstruction, pushing the boundaries of the state of the art and opening up new avenues for progress in this crucial research domain.


\section{Related Work} 
\paragraph{Implicit Method} 
Traditional implicit methods include poisson surface reconstruction\cite{2013Poisson}. Then IPSR\cite{ipsr} remove the restriction of normal by introducing a iterative method to guess the unoriented normal.
Implicit methods, which are mostly deep methods~\cite{2019DeepSDF, 2019Occupancy, 2020Convolutional, 2020SSRNet, 2022POCO, SAL, Alto, galerkin}, have gained significant attention in the field of point cloud surface reconstruction. With advances in deep learning, researchers have focused on leveraging the strong representation ability of neural networks to build better surfaces from point clouds. Neural networks, being able to theoretically fit any function, have been utilized in various ways to learn the distance function or occupancy value of shapes. 

One popular approach, DeepSDF~\cite{2019DeepSDF}, encodes the entire shape as a coding and then decodes the signed distance function (SDF) of any query point based on the coding and its position. Occupancy Networks~\cite{2019Occupancy}, on the other hand, predict only the occupancy value of query points, disregarding distance prediction. It extracts features from the point cloud and uses these features along with position to predict the occupancy value. Convolutional Occupancy Network~\cite{2020Convolutional} builds upon Occupancy Networks~\cite{2019Occupancy} by integrating 3D convolution into the feature learning process and applying interpolation to obtain the feature of the query point. Chen\cite{2023Unsupervised} propose an unsupervised method to learn SDF and reconstruct the mesh.

Points2Surf~\cite{2020Points2Surf} aims to regress the absolute SDF value based on local information and classify the sign based on global information. POCO~\cite{2022POCO} preserves the features of the input point cloud and incorporates a learning-based interpolation module to achieve high-performance reconstruction. NDF~\cite{2020ndf} introduces the concept of unsigned distance function to represent open surfaces. Neural Pull~\cite{2020npull} represents a 3D shape using a single network model and trains the network by predicting the shift of query points. 

On-surface Prior~\cite{onprior} adds a geometry prior to Neural Pull, enforcing that the norm of the query point shift to the surface should be minimal. CAP~\cite{capudf} extends the SDF representation of Neural Pull to unsigned distance functions (UDF) and achieves impressive performance. Dynamic Code~\cite{2022dynamicCode} proposes combining local code with local position encoding to optimize the position of the code. MeshUDF\cite{2021MeshUDF} tries to extract mesh directly from the UDF. Neural-imls\cite{neuralimls2023wang} combines moving least sqaure and implicit representation to represent local gemoetry. NeuralIndicator\cite{NIndicator} tries to learn the implicit function based on neural indicator priors.

From a network design perspective, these methods can be categorized into global methods (e.g., DeepSDF), local methods (e.g., Occupancy Networks, Convolutional Occupancy Network, POCO), and their combinations (e.g., Points2Surf). From the representation method standpoint, these methods can be divided into SDF-based methods (e.g., DeepSDF, Occupancy Networks, Convolutional Occupancy Network, Points2Surf, POCO, Neural Pull, Dynamic Code) and UDF-based methods (e.g., NDF, On-surface Prior, CAP).

\paragraph{Explicit Method}

Explicit methods, such as mesh-based methods, spline functions, and moving least squares (MLS)~\cite{mls}, have been extensively studied in surface reconstruction for several decades. These methods involve constructing a local geometric model, such as a plane, curve, or geometric primitive, to fit the local surface. In our work, we focus on the moving least squares (MLS) family of methods, which includes several relevant approaches that we will discuss in detail.

Moving least squares (MLS) is a popular explicit method that establishes a local coordinate system for each patch and fits a curve based on the local patch information. Implicit moving least squares (IMLS)~\cite{imls} is an algebraic method that simplifies the local geometry by approximating the curve with a plane. It then merges the local planes using adaptive weights. Robust moving least squares (RMLS)~\cite{rmls} extends MLS by establishing the neighborhood using a forward-search paradigm, enabling it to handle noisy input and capture sharp features.

However, these explicit methods often struggle to leverage the capabilities of neural networks to improve surface reconstruction. They rely on predefined geometric models and may not effectively capture complex or irregular surfaces.

\paragraph{Combination of Implicit and Explicit Method} 

DeepMLS~\cite{deepmls} is a successful demonstration of the fusion of explicit and neural network-based methods, building upon the Implicit Moving Least Squares (IMLS)~\cite{imls} technique. The DeepMLS approach initiates by utilizing a neural network to predict the parameters of local planes. These local planes are subsequently merged using an adaptive weight scheme, resulting in an implicitly represented surface through the calculation of the Signed Distance Function (SDF).
While DeepMLS capitalizes on the explicit IMLS method to facilitate the calculation, our work further combines the implicit and explicit methods. It is crucial to discuss these divergent approaches. The implicit method can be readily integrated with neural networks, but it may sacrifice the geometric prior information. Conversely, explicit methods necessitate specific designs to synergize with neural networks, and the flexibility of the geometric prior can be restricted in certain scenarios.
In our research, our objective is to enhance the flexibility of the geometric prior compared to the baseline methods, IMLS~\cite{imls} and DeepMLS~\cite{deepmls}. By amalgamating components from both implicit and explicit methods, we strive to devise a novel approach that leverages the strengths of each approach while mitigating their respective limitations.

\section{Method}
\label{sec:method}
\begin{figure*} 
\centering
  \includegraphics[width=2\columnwidth]{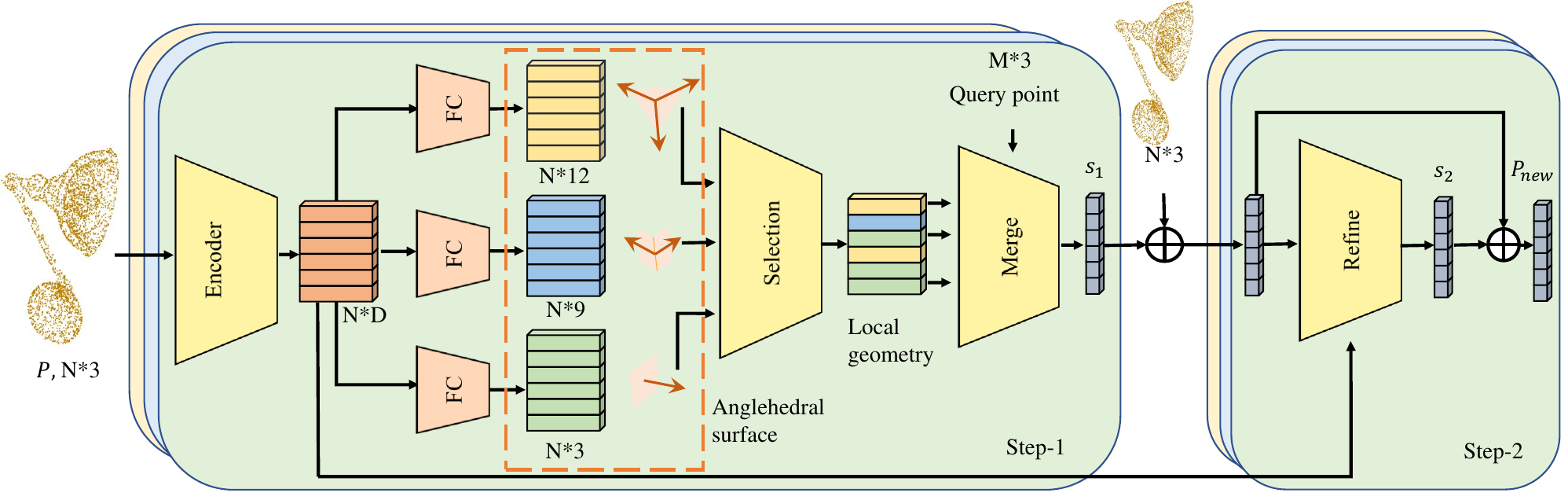}
  \caption{The pipeline of the whole processing.}
  \label{fig:pipeline}
\end{figure*}

In this section, we present an approach of local surface representation for point cloud reconstruction, referred to as anglehedral surface. Fig.~\ref{fig:pipeline} provides an overview of our proposed pipeline. Firstly, we outline the problem and introduce the pipeline. Secondly, we describe the anglehedral surface representation and the methodology for computing the distance from the query point to the anglehedral surface. Thirdly, we discuss the selection of the local surface representation and the merging process for these representations. Fourthly, we elaborate on the refinement method. Lastly, we address the training loss in our approach.

\subsection{Main Pipeline}
\paragraph{Formalization} First, we formalize our method. We define the space as $\mathcal{S}=[-1,1]^3$. Given an input point cloud $\mathcal{P}=\left\{p_1, p_2, \cdots, p_n, p_i \in \mathcal{S}\right\}$, our objective is to learn a function represented by a neural network $s(q)=\mathcal{F}(\mathcal{P}, q)$, where $q \in \mathcal{S}$. The output of the function $\mathcal{F}$ represents the shortest displacement from the point $q$ to the implicit surface, denoted as $s(q) \in \mathcal{R}^3$. In contrast to the completely implicit method, we introduce a geometric prior during the UDF prediction process, which will be explained in detail. Moving forward, we need to extract the surface from the function $\mathcal{F}$. This surface corresponds to the zero-level set of the displacement, defined as $\mathcal{M}=\left\{q \mid\|s(q)\|_2=0, q \in \mathcal{S}\right\}$.

\paragraph{Explanation of Output} 
The direct output of neural network is $s(q)$, where $q$ is the query point. As per the definition of $s(q)$, it is evident that $\|s(q)\|_2$ represents the UDF (Unsigned Distance Function) of $q$. By defining the output in this way, we can achieve self-supervised learning of UDF prediction. Although numerous point cloud reconstruction techniques, such as~\cite{2022POCO, 2020Convolutional, deepmls, gifs}, necessitate additional surface information for supervision, like the SDF (Signed Distance Function) of the query point or determining the relative positions of two query points with respect to the surface. Our approach does not depend on any additional supervision beyond the input point cloud. This is achieved because we can readily generate a new point cloud using the function $\mathcal{F}$. Specifically, the new point cloud is derived as $\mathcal{P}_{\text {new }}=\{q+s(q) \mid q \in \mathcal{S}\}$. By employing the chamfer distance, we aim to bring the two point clouds, $\mathcal{P}_{\text {new }}$ and $\mathcal{P}$, closer to each other. This methodology streamlines the training process by facilitating the generation of a point cloud from $s(q)$. Conversely, if the output of $\mathcal{F}$ were the UDF of the query point, determining the displacement by calculating the gradient of the query point would be notably time-consuming.

\paragraph{Architecture}
The main architecture is illustrated in Fig. \ref{fig:pipeline}. The backbone of the architecture, referred to as the encoder, comprises 4 layers of point cloud transformers~\cite{zhao2021pointtrans} responsible for extracting per-point features from the point cloud. Each point $p_i$ within the point cloud is associated with its corresponding per-point feature $f_i$. The subsequent step involves predicting the local geometry based on these features. In the case of the local geometry being a plane, we predict the normal of the plane. If it represents an anglehedral surface, we predict the associated normals. The determination of the local geometry type depends on the minimum average distance from neighboring points to the surface of the local geometry. This approach enhances the flexibility of the geometry prior in contrast to solely considering planes as in SuperUDF~\cite{superudf}. Furthermore, the necessity of a local coordinate system is eliminated. 

After obtaining the per-point features and local surface parameters, we employ a two-step approach to calculate the displacement from a query point to the implicit surface. In the first step, for each query point, we identify the $k_2$ nearest neighbors in the point cloud and compute the displacement from the query point to each neighbor's surface. These displacements are then averaged using weights generated by a learning-based weight generator. In the second step, a neural network is utilized to directly derive a refined displacement from the shifted query point to the target surface. The two displacements are then combined to yield the final displacement. By executing these two sequential steps during the decoding phase, we are able to determine the displacement from any query point in space to the nearest surface.

The architecture is composed of the encoder backbone and the decoder part, which together form the function $\mathcal{F}(\mathcal{P}, q)$.

\subsection{Plane}
We introduce three local geometric models to improve the adaptability of the local geometry prior in scenarios where a local coordinate system is not accessible. The first model is a plane. When dealing with flat surfaces, a plane serves as a suitable representation for the local patch. Defining a plane only requires estimation of the normal vector and a central point. More precisely, for a point $p_i$ within the point cloud, we employ three FC (fully connected layers) to predict the normal vector in the following manner:
\begin{equation}
\begin{aligned}
    n_i = MLP_1(f_i) \\
    n_i := \frac{n_i}   {\|n_i\|_2},
\end{aligned}
\label{eq:normal1}
\end{equation}
Once the plane has been defined, we can effortlessly compute the displacement from any query point $q_j$ to the plane using the following equation:

\begin{align}
    s_{plane}(p_i, q_j) = \langle p_i - q_j, n_i \rangle n_i
\end{align}

Here, $\langle \cdot, \cdot \rangle$ denotes the inner product.

\subsection{Dihedral Surface}
\begin{figure}
\centering
  \includegraphics[width=\columnwidth]{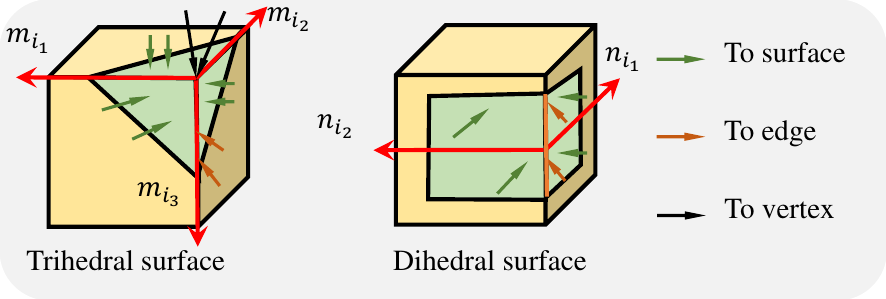}
  \caption{The distance from query point to trihedral surface and dihedral surface.}
  \label{fig:distance}
\end{figure}

The dihedral surface is a versatile structure that can accommodate a variety of shapes, including patches in tables, chairs, and more. It offers a distinct advantage over merging multiple planes by enabling the preservation of sharp features on the curved surface, as illustrated in Fig. \ref{fig:distance}. Describing the dihedral surface necessitates the use of two normals. These normals, akin to planes, are derived from the characteristics of point $p_i$ through the utilization of a MLP (Multi-Layer Perceptron).
\begin{equation}
\begin{aligned}
    n_{i_1}, n_{i_2}, c_i &= MLP_2(f_i), \\
    n_{i_1} &:= \frac{n_{i_1}}{\|n_{i_1}\|_2},\\
    n_{i_2} &:= \frac{n_{i_2}}{\|n_{i_2}\|_2},\\
    c_i &= Tanh(c_i)*0.01
\end{aligned}
\label{eq:normal2}
\end{equation}
Where $n_{i_1}$ and $n_{i_2}$ represent the normals, $c_i$ denotes a local displacement that can adjust the dihedral surface. The necessity for a local displacement is apparent: $p_i$ may not align precisely with the edge of the turning surface, and a local displacement can accurately position the dihedral surface.

In order to compute the distance from a query point $q_j$ to the dihedral surface of $p_i$, it is essential to identify the nearest point on the dihedral surface to $q_j$. This procedure is delineated in Algorithm \ref{alg:s2fold}, with a visual depiction provided in Fig. \ref{fig:distance} for reference.

\begin{algorithm}
\DontPrintSemicolon
  \SetAlgoLined
  \KwIn {$n_{i_1},n_{i_2},c_i,p_i,q_j$}
  \KwOut {$s_{dihedral}(p_i, q_j)$}
  initialization\;
  $s_{dihedral}(p_i, q_j)=\inf$\;
  $s_{i_1}=D(q_j, halfplane_{i_1})$\tcc*[f] {the shortest displacement from $q_j$ to half plane controlled by $n_{i_1}$}\;
  $s_{i_2}=D(q_j, halfplane_{i_2})$\tcc*[f] {the shortest displacement from $q_j$ to half plane controlled by $n_{i_2}$}\;
  $s_{dihedral}(p_i, q_j)=min(s_{i_1}, s_{i_2})$ \tcc*[f] {min means vector with minimum $L_2$ norm}\;
  \caption{Calculate $s_{dihedral}(p_i, q_j)$}
  \label{alg:s2fold}
\end{algorithm}

\subsection{Trihedral Surface}
We employ three normals to characterize the trihedral surface, as illustrated in Fig.\ref{fig:distance}. This type of surface is versatile, capable of representing the sharp edges of a cube corner and the conical shape of a cone. Just like the dihedral surface, utilizing the trihedral surface to depict corners or cones maintains the integrity of sharp features. The estimation of the three normals for the trihedral surface is derived from the feature $f_i$ of point $p_i$ through the following equations:

\begin{align}
    m_{i_1}, m_{i_2}, m_{i_3}, d_i& = MLP_3(f_i),\\
    m_{i_1} &:= m_{i_1}/\|m_{i_1}\|_2,\\
    m_{i_2} &:= m_{i_2}/\|m_{i_2}\|_2, \\
    m_{i_3} &:= m_{i_3}/\|m_{i_3}\|_2, \\
    d_i &= Tanh(d_i)*0.01,
\end{align}
where $m_{i_1}, m_{i_2}, m_{i_3}$ are the normals and $d_i$ is the local displacement of the trihedral surface center.

In order to compute the distance from the query point $q_j$ to the trihedral surface of $p_i$, we must identify the nearest point on the trihedral surface to $q_j$. This process is outlined in Algorithm \ref{alg:s3fold}, and a visual depiction can be found in Fig~/ref{fig:distance}.

\begin{algorithm}
\DontPrintSemicolon
  \SetAlgoLined
  \KwIn {$m_{i_1},m_{i_2},m_{i_3},c_i,p_i,q_j$}
  \KwOut {$s_{trihedral}(p_i, q_j)$}
  initialization\;
  $s_{trihedral}(p_i, q_j)=\inf$\;
  $s_{i_1}=D(q_j, halfplane_{i_1})$\tcc*[f] {the shortest displacement from $q_j$ to half plane controlled by $m_{i_1}$}\;
  $s_{i_2}=D(q_j, halfplane_{i_2})$\tcc*[f] {the shortest displacement from $q_j$ to half plane controlled by $m_{i_2}$}\;
  $s_{i_3}=D(q_j, halfplane_{i_3})$\tcc*[f] {the shortest displacement from $q_j$ to half plane controlled by $m_{i_3}$}\;
  $s_{trihedral}(p_i, q_j)=min(s_{i_1}, s_{i_2},  s_{i_3})$ \tcc*[f] {min means vector with minimum $L_2$ norm}\;
  \caption{Calculate $s_{trihedral}(p_i, q_j)$}
  \label{alg:s3fold}
\end{algorithm}

\subsection{Choose Method}
After constructing three local geometries, the next crucial step is to determine the most suitable one for point $p_i$. We have developed a straightforward yet efficient method for making this selection. The fundamental idea is that when a correct local geometry is created, all adjacent points should precisely align with the surface. To elaborate, given a specific point $p_i$, we choose $k_1$ neighboring points from the point cloud $\mathcal{P}$, represented as $\{p_{i_1},\cdots,p_{i_{k_1}}\}$. Subsequently, we compute the average distance for each local geometry using the following formulas:

\begin{align}
    D_{plane_i} = \frac{1}{k_1}\sum_{l=0}^{k_1} s_{plane}(p_i, p_{i_l}) \\
    D_{dihedral_i} = \frac{1}{k_1}\sum_{l=0}^{k_1} s_{dihedral}(p_i, p_{i_l})\\
    D_{trihedral_i} = \frac{1}{k_1}\sum_{l=0}^{k_1} s_{trihedral}(p_i, p_{i_l})\\
    D_i = \min(D_{plane_i}, D_{dihedral_i},D_{trihedral_i}),
\end{align}
where $min()$ is a function that selects the vector with the minimum $L_2$ norm, the local geometry that results in the minimum average distance, determined by $I_i = \text{argmin}(D_{\text{plane}_i}, D_{\text{dihedral}_i}, D_{\text{trihedral}_i})$, is selected as the final method.

\subsection{Local Geometry Merge}
Through the aforementioned pipeline, we can construct the local geometry for each point $p_i$. However, these individual local geometries must be combined using adaptive weights. Instead of directly merging the local geometries, we calculate the average displacement for each point. Specifically, for a query point $q_j$, we first identify its $k_2$ neighboring points in $\mathcal{P}$, denoted as $\{p_{j_1}, \cdots, p_{j_{k_2}}\}$. Each neighboring point has its own local geometry and corresponding displacement, defined as $s_{g}(p_{j_l}, q_j) = (s_{\text{plane}}(p_{j_l}, q_j), s_{\text{dihedral}}(p_{j_l}, q_j), s_{\text{trihedral}}(p_{j_l}, q_j))[I_{j_l}]$, where $[I_{j_l}]$ denotes the index. Subsequently, the averaged displacement based on the local geometries is computed as follows:
\begin{equation}
    s_{g}(q_j) = \sum_{l=0}^{k_2} w(p_{j_l}, q_j) \cdot s_{g}(p_{j_l}, q_j),
\end{equation}
Here, $w(p_{j_l}, q_j)$ represents the adaptive weight learned through a neural network.

Regarding the weight $w(p_{j_l}, q_j)$, we first predict a scale parameter for each point $p_i$ based on its point feature $f_i$ using a MLP with four layers:
\begin{equation}
    r_i = MLP_4(f_i).
\end{equation}

Then, we consider the relative distance and the scale parameter together using the Softmax operator,
\begin{equation}
\begin{aligned}
    w(p_{j_1}, q_j), \cdots, w(p_{j_{k_2}}, q_j)=\\
    Softmax(-\frac{\|p_{j_1}-q_j\|_2}{\theta r_i}, \cdots, -\frac{\|p_{j_{k_2}}-q_j\|_2}{\theta r_i}),
\end{aligned}
\label{eq:merge_weight}
\end{equation}
where $\theta$ is a hyper-parameter.

\subsection{Displacement Refinement}
After the initial data processing in the aforementioned pipeline, the query point is displaced to the surface in accordance with the local geometry. Subsequently, a basic neural network is utilized to enhance the displacement accuracy. To elaborate, for the displaced query point $q_j+s_{g}(q_j)$, the local geometry is encoded using positional encoding, which is then fed into a MLP. This process can be represented as:

\begin{equation}
    PE(q_j+s_{g}(q_j), p_i) = MLP_5(Positional(q_j+s_{g}(q_j)-p_i)),
\end{equation}

Here, the function $Positional$ represents positional encoding~\cite{2020Fourier}. Subsequently, a point transformer~\cite{zhao2021pointtrans} is utilized to obtain the feature of $q_j$, as shown in the following equation:

\begin{equation}
    f_{q_j+s_{g}(q_j)} = \sum_{l=1}^{k_2}PointTrans(PE(q_j+s_{g}(q_j), p_{j_l}), f_{j_l}).
\end{equation}

Finally, a MLP followed by a Tanh activation function is employed to predict the refined displacement:

\begin{equation}
    s_{refined}(q_j+s_{g}(q_j)) = Tanh(MLP_6(f_{q_j+s_{g}(q_j)})).
\end{equation}

Consequently, the total displacement is determined by the sum of the geometric displacement $s_{g}(q_j)$ and the refined displacement $s_{refined}(q_j+s_{g}(q_j))$, represented as $s(q_j) = s_{g}(q_j) + s_{refined}(q_j+s_{g}(q_j))$.

\subsection{Losses} 
Our training process is self-supervised, with the sole source of supervision being the input point cloud. In this training framework, we incorporate three distinct loss functions. Firstly, our objective is for $\mathcal{F}(\mathcal{P}, q)$ to accurately predict the displacement from any query point to the nearest surface. To accomplish this, we generate a new point cloud, $\mathcal{P}_{new}$, based on the output of the function $\mathcal{F}(\mathcal{P}, q)$. The sampling technique is straightforward: a set of $m$ points is randomly selected around the original point cloud $\mathcal{P}$ to form the query point set $Q={q_1, \cdots, q_m}$. Subsequently, the displacement of each query point $s(q_i)$, where $i \in \{1, \cdots, m\}$, is calculated. Finally, the new point cloud set $\mathcal{P}_{new}$ is defined as $\{q_1+s(q_1), \cdots, q_m+s(q_m)\}$. The distance between $\mathcal{P}$ and $\mathcal{P}_{new}$ is measured using the chamfer distance, which is expressed as:

\begin{align}
L_{CD_1}=\frac{1}{m} \sum_{j=1}^{m}\min_{i=1}^{n}\left\|q_j+s(q_j)-p_i\right\|_1.
\label{eq:cd_1}
\end{align}

\begin{figure*} 
\centering
  \includegraphics[width=1.8\columnwidth]{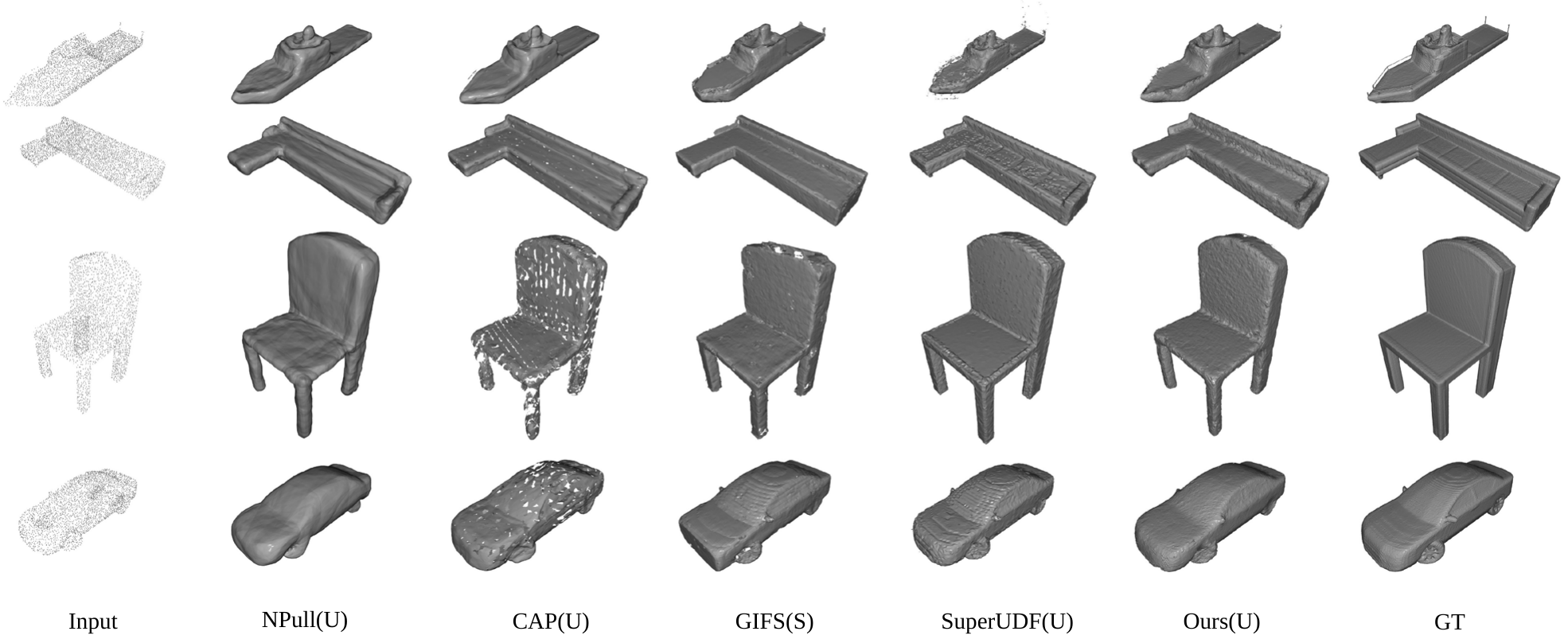}
  \caption{Visualized results of our method and state-of-the-art alternatives on ShapeNet. (S) mean supervised method and (U) means unsupervised method. Methods are NPull, CAP, GIFS, SuperUDF and Ours. The number of input point cloud is 3000.}
  \label{fig:shapenet_result1}
\end{figure*}
To enhance the network's ability to predict accurate local geometric parameters, such as the normal of a plane and the normals of anglehedral surfaces, we introduce a local displacement regularization term. The core concept is to reduce the displacement between query points and target surfaces. For each point $p_i$ within the point cloud, we identify its $k_1$-nearest neighbors in the same cloud $\mathcal{P}$. These $k_1$ neighboring points are expected to be situated on the local geometry surrounding point $p_i$. The second loss function is then formulated as follows:
\begin{align}
    L_{local} = \frac{1}{n}\sum_{i=1}^{n}D_i.
    \label{eq:local}
\end{align}

Third, we incorporate a UDF regularization term, as proposed in~\cite{superudf}, to ensure a continuous UDF distribution. The main idea is simple: the UDF should decrease evenly with a gradient of 1 from the query point to the surface, and the gradient direction should not change. This can be deduced from the definition of the UDF. For a query point $q$, its UDF is $\|s(q)\|_2$. For any point $\Tilde{q}$ in the path from $q$ to the surface, the displacement direction should be the same as that of $q$, and the UDF should decrease with a gradient of 1. Let $\Tilde{q} = q + (1-\alpha)s(q)$, where $\alpha \in [0, 1]$. The corresponding regularization term is given by:
\begin{align}
    L_{UDF} = \|\frac{s(\Tilde{q})}{1-\alpha}-s(q)\|_2.
\end{align}


\section{Experiments}\label{4}

In this section, we evaluate the performance of our method on the 3 datates, including ShapeNetCore ~\cite{chang2015shapenet}, ABC~\cite{abc} and ScanNet~\cite{dai2017scannet}. We first show quantitative and qualitative comparison with previous state-of-the-art methods. Next, We conduct comprehensive ablation studies and robust analysis. Quantatively, we choose $CD_1$ to measure the completeness and accuracy of reconstructed mesh. 
In short, we show how we make use of chamfer distance to constrain the training. $CD_1$ equation is shown as the Eq.(\ref{eq:chamfer}),
\begin{equation}
\label{eq:chamfer}
\begin{aligned}
\mathrm{CD}_{1}=& \frac{1}{2 N_{x}} \sum_{i=1}^{N_{x}}\left\|\mathbf{x}_{i}-\mathcal{S}_{y }\left(\mathbf{x}_{i}\right)\right\|_1+\\
& \frac{1}{2 N_{y}} \sum_{i=1}^{N_{y}}\left\|\mathbf{y}_{i}-\mathcal{S}_{x}\left(\mathbf{y}_{i}\right)\right\|_1,
\end{aligned}
\end{equation}
$\{ \mathbf{x}_{i}, i=0\cdots N_x\}$ is the points set sampled from ground-truth mesh. $\{\mathbf{y}_{i}, i=0\cdots N_y\}$ is the point set sampled from reconstructed mesh. $\mathcal{S}_{y}(\mathbf{x}_{i})$ means the nearest point to $\mathbf{x}_{i}$ in reconstructed point set. $\mathcal{S}_{x}(\mathbf{y}_{i})$ means the nearest point to $\mathbf{y}_{i}$ in ground-truth point set. $\|\cdot\|_1$ means $L1-distance$.

%

\subsection{Implementation Details}
\label{sec:implementation}

\paragraph{Network architecture} Our network backbone consists of four Point Transformer blocks, with the dimensions of $64$, $128$, $256$, $256$ respectively. We use $k=36$ nearest neighbor points as the local patch. After obtaining per-point feature, the network has two branches. In the first branch, we use the neighbor points of input point cloud to supervise the trihedral surface parameter, the number of neighbor points $k_1$ is $36$. In another branch, we predict $r_i$ in Eq. (\ref{eq:merge_weight}) of every point. After merging the local anglehedral surface, we adopt a refinement layer implemented by point transformer, the k-nearest neighbor of the refinement step is $12$ and the feature dimension is $64$. Finally, we adopt a MLP to predict the 3-dimension refinement shift. 
During training, we adopt cosine learning rate adjustment~\cite{2016coslr}, the initial learning rate is 0.001. We use Adam~\cite{adam} optimizer to learn the parameter. The batch-size is 2. The query point number is 6000. The query point for training is obtained by randomly sampling point around the original point with the uniform distribution $U(-0.03, 0.03)$. During testing, to obtain the UDF in the whole space, we first partition the space into $256^3$ voxels. We use the network to predict the UDF of the voxel center. Because our representation is not SDF, we do not use Marching Cube\cite{lorensen1987marching} to extract mesh. After obtaining the UDF of every voxel, we apply this~\cite{superudf} to extract the mesh.

\begin{figure*} 
\centering
  \includegraphics[width=1.8\columnwidth]{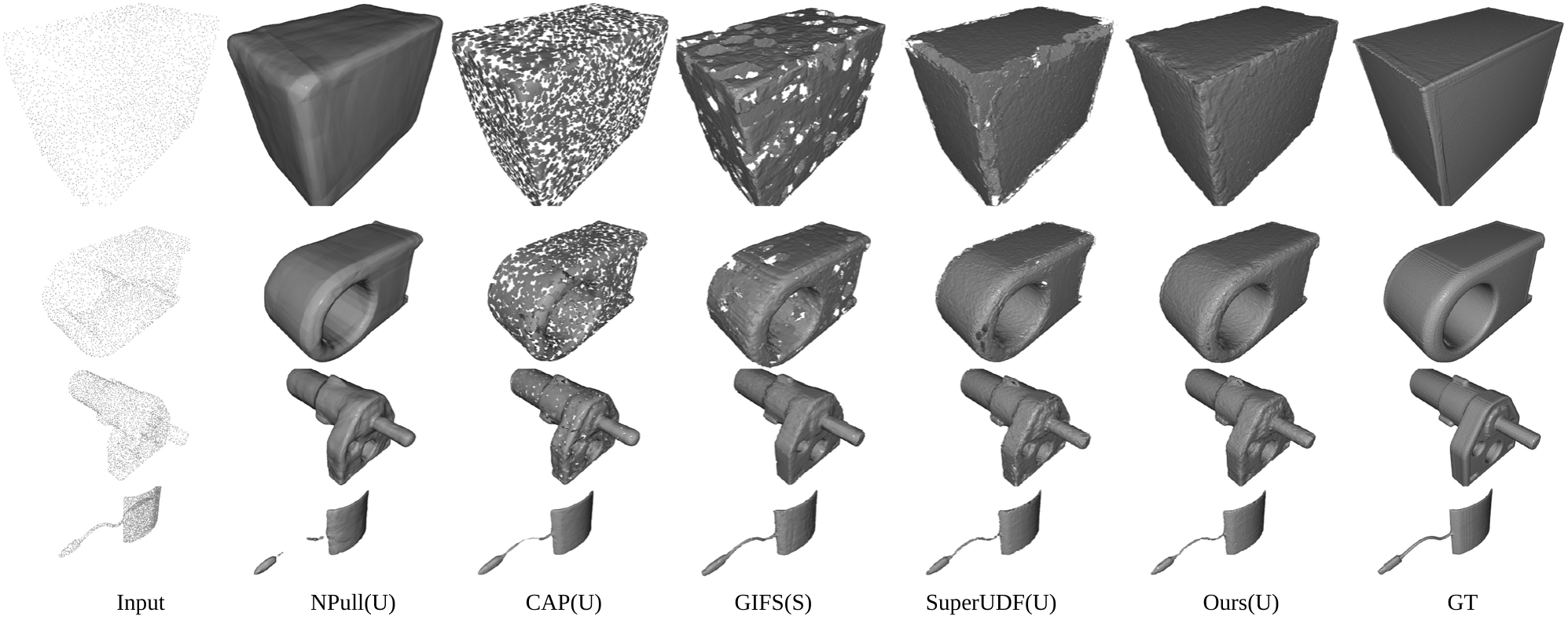}
  \caption{Visualized results of our method and state-of-the-art alternatives on ABC 001. (S) mean supervised method and (U) means unsupervised method. Methods are NPull, CAP, GIFS, SuperUDF and Ours. The number of input point cloud is 3000.}
  \label{fig:abc_result2}
\end{figure*}
\subsection{Shape Surface Reconstruction}
In order to show the reconstruction performance of anglehedral surface on general object without much sharp feature, we first select a most widely-used dataset for point cloud reconstruction, ShapeNet. It contains 13 classes. For every class, we random select 100 shapes for test. Because the whole pipeline is self-supervised, we do not need to partition the train and test dataset. The result is shown in Tab.~\ref{tab:shapenet}. From the Tab.~\ref{tab:shapenet}, we can see that in most classes, our method outperform other method. Especially, we should focus on the comparison between ours and SuperUDF~\cite{superudf}. The most important difference between ours and SuperUDF~\cite{superudf} is the geometry prior. We use anglehedral surface while SuperUDF use a simple plane instead. Thus, from the comparison, we can see the improvement from the anglehedral surface. We can see that ours can outperform SuperUDF in most classes. 

\begin{table}[]
    \centering
\begin{tabular}{l|lllll}
\toprule
Class     & GIFS            & CAP    & NPull & SuperUDF & Ours            \\

\midrule
airplane  & 0.0026          & 0.0031 & 0.0037 & \textbf{0.0022}  & 0.0023 \\
bench     & 0.0058          & 0.0059 & 0.0079 & 0.0029  & \textbf{0.0027} \\
cabinet   & 0.0104          & 0.0048 & 0.0067 & 0.0039  & \textbf{0.0036} \\
car       & 0.0052          & 0.0054 & 0.0060 & 0.0029  & \textbf{0.0025} \\
chair     & 0.0041          & 0.0042 & 0.0122 & 0.0039  & \textbf{0.0038} \\
display   & 0.0055          & 0.0047 & 0.0049 & \textbf{0.0034}  & 0.0041 \\
lamp      & 0.0078          & 0.0082 & 0.0089 & 0.0030  & \textbf{0.0027} \\
speaker   & 0.0081          & 0.0063 & 0.0075 & 0.0041  & \textbf{0.0038} \\
rifle     & 0.0029 & 0.0013 & 0.0016 & 0.0011  & \textbf{0.0010} \\
sofa      & 0.0069 & 0.0050 & 0.0051 & 0.0033  & \textbf{0.0030} \\
table     & 0.0065          & 0.0088 & 0.0117 & 0.0025  & \textbf{0.0022} \\
phone & 0.0049          & 0.0024 & 0.0028 & 0.0021  & \textbf{0.0019} \\
vessel    & 0.0033          & 0.0025 & 0.0037 & 0.0021  & \textbf{0.0021} \\
\midrule
Mean      & 0.0057          & 0.0048 & 0.0063 & 0.0028  & \textbf{0.0027}\\
\bottomrule
\end{tabular}
    \caption{$CD_1$ comparison on ShapeNet within 13 classes}
    \label{tab:shapenet}
\end{table}
\begin{figure*} 
\centering
  \includegraphics[width=1.8\columnwidth]{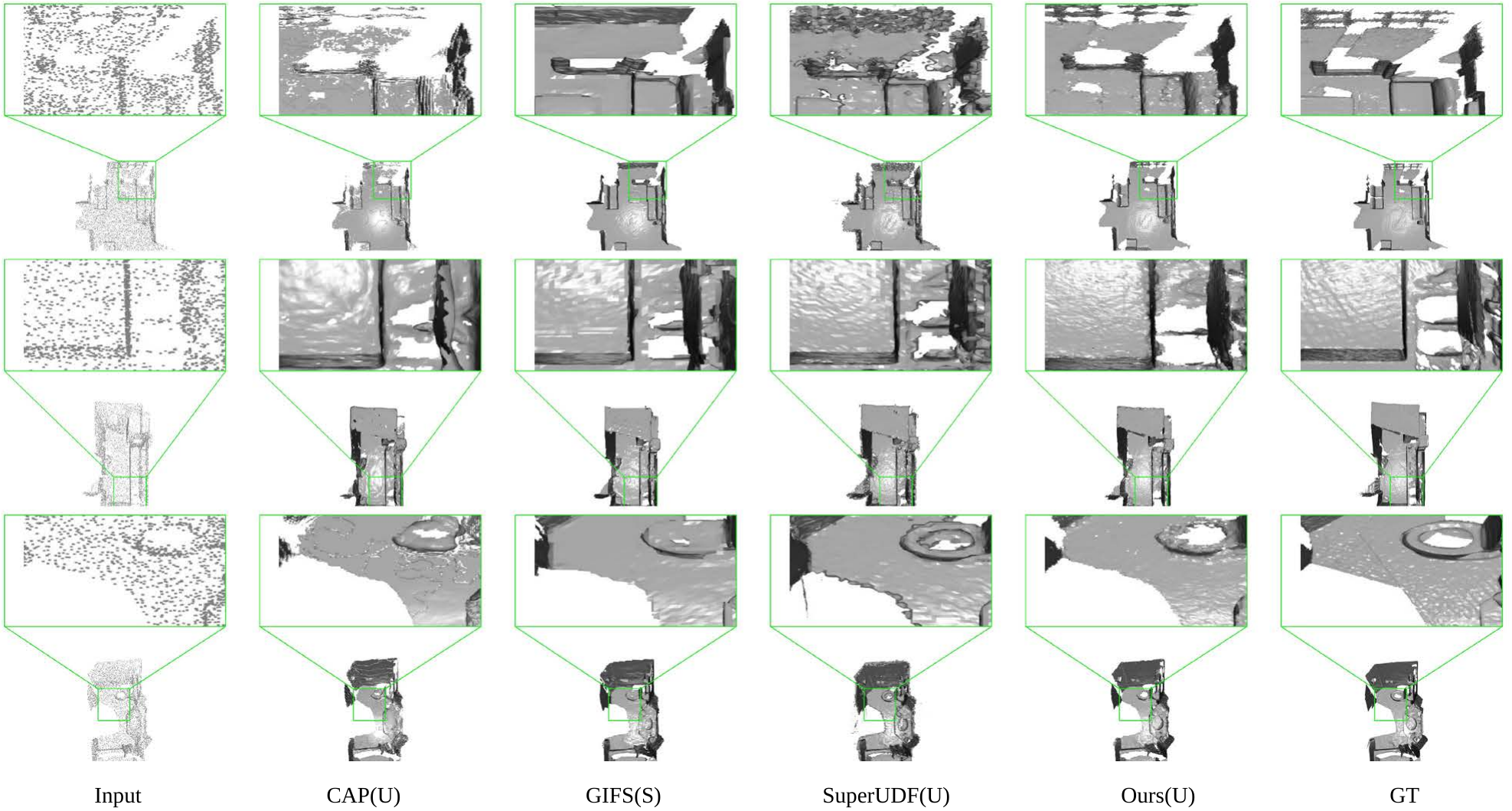}
  \caption{Visualized results of our method and state-of-the-art alternatives on ScanNet. (S) mean supervised method and (U) means unsupervised method. Methods are CAP, GIFS, SuperUDF and Ours. The number of input point cloud is 10000.}
  \label{fig:scan_result1}
\end{figure*}
Further more, we provide the visualization result for comparison as well. As shown in Fig.~\ref{fig:shapenet_result1}, we can see that ours can generate mesh faithful to the ground-truth.  

Next, we show our result on ABC~\cite{abc} 001. There are over 7000 shapes in ABC 001. Similar to ShapeNet, we randomly sample 100 shapes from it. The reason why we choose ABC is that shapes in ABC have lots of sharp feature. The result is shown in Tab.~\ref{tab:abc}. We can see that our method outperforms all other method. What's more, we provide the visualization result in Fig.~\ref{fig:abc_result2}. There is an important problem in UDF-based reconstruction. If UDF is not correct or noisy near the surface, the reconstructed mesh might have some holes. By comparing the ours and others, we can see that other methods have lots of holes except NPull~\cite{2020npull}, especially near the turning face and our method can generate mesh with less numer of holes, while NPull is SDF-based method and has its own problem over-smooth. Because all the UDF-based methods mentioned above use the same mesh extraction method~\cite{superudf} and our method can generate the mesh with the least number of holes, anglehedral surface can improve the UDF quality especially near the turning surface.

\begin{table}[]
\centering
\begin{tabular}{l|lllll}
\toprule
Method&SuperUDF      & CAP      & NPull     & GIFS      & Ours     \\

\midrule
$CD_1$&0.0063 & 0.0077 & 0.0092 & 0.0075 & \textbf{0.0051} \\
\bottomrule
\end{tabular}
\caption{$CD_1$ comparison results on ABC 001 dataset.}
\label{tab:abc}
\end{table}

\subsection{Scene Surface Reconstruction}
In order to show the performance on open surface, we select the partially scanned scene, ScanNet~\cite{dai2017scannet}. First, we provide the quantitative result in Tab.~\ref{tab:scan}. We can see that our method can outperform other methods. What's more, we provide the visualization result on final mesh in Fig.~\ref{fig:scan_result1}. Comparing with SuperUDF~\cite{superudf}, our method can generate boundary faithful to the GT, while the boundary of SuperUDF expands outward a lot. This is for the reason why the local geometry prior plane near the boundary will expand and cannot fit the local half plane. However, our anglehedral surface can fit the half plane thus can generate better boundary. 

\begin{table}[]
\centering
\begin{tabular}{l|llll}
\toprule
Method&SuperUDF& CAP&GIFS & Ours     \\
\midrule
 $CD_1$&0.0039 & 0.0044 & 0.0043 & \textbf{0.0033} \\
\bottomrule
\end{tabular}
\caption{$CD_1$ comparison on ScanNet with the number of sample point 10000.}
\label{tab:scan}
\end{table}

\subsection{Results on different Input}
Although our method cannot generate good performance on sparse and noisy input, we list the result in Tab.~\ref{tab:sparse} and Tab.~\ref{tab:noise}. When introducing noise, we add the input point cloud and the Gaussian noise with mean value as 0 and different $\sigma$.
\begin{table}[]
\centering
\begin{tabular}{l|llll}
\toprule
\#points &500      & 1000      & 3000     & 6000\\

\midrule
$CD_1$&0.0143 & 0.0089 & 0.0051  & 0.049 \\
\bottomrule
\end{tabular}
\caption{$CD_1$ comparison results on ABC 001 dataset.}
\label{tab:sparse}
\end{table}

\begin{table}[]
\centering
\begin{tabular}{l|llll}
\toprule
$\sigma$ &0.000      & 0.001      & 0.002     & 0.003\\

\midrule
$CD_1$&0.0051 & 0.0059 & 0.0081  &0.093  \\
\bottomrule
\end{tabular}
\caption{$CD_1$ comparison results on ABC 001 dataset.}
\label{tab:noise}
\end{table}
\subsection{Ablation Study}
\begin{figure*} 
\centering
  \includegraphics[width=1.8\columnwidth]{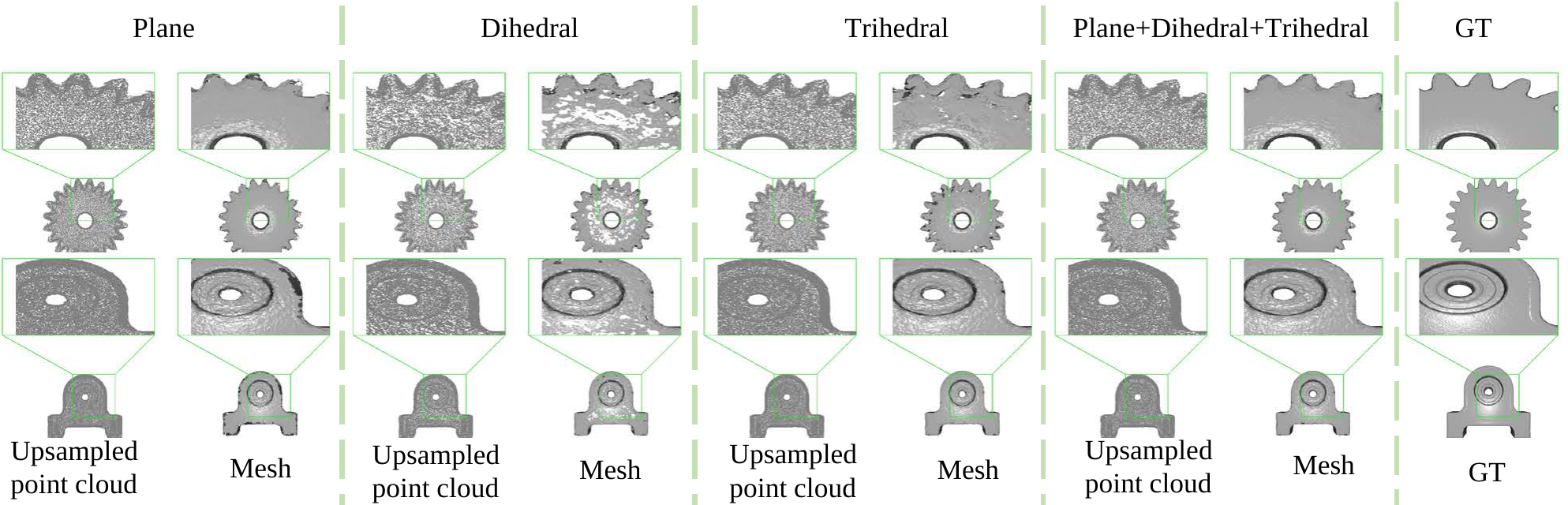}
  \caption{Evaluation of the effectiveness of the anglehedral surface on flat surface and turning surface.}
  \label{fig:ablation_abc}
\end{figure*}

\begin{figure*} 
\centering
  \includegraphics[width=1.8\columnwidth]{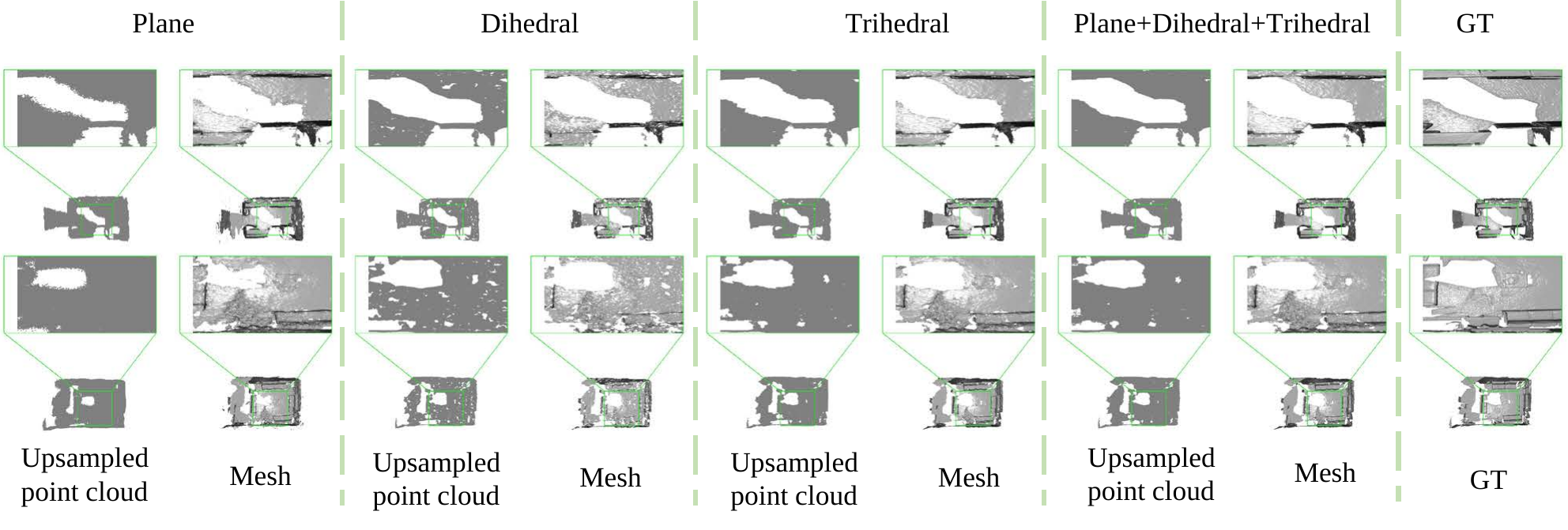}
  \caption{Evaluation of the effectiveness of anglehedral surface on the boundary of open surface.}
  \label{fig:ablation_scan}
\end{figure*}

The key contribution of this paper is the anglehedral surface, including dihedral and trihedral surface as the geometry prior. Therefore, we want to show how much contribution every part makes. In total, we design 4 experiments. In detail, we show the result under different compositions of the 3 geometry prior in Tab.~\ref{tab:ablation}. We are not surprised to see that plane+dihedral+trihedral achieves best result. 
\begin{table}[]
\centering
\begin{tabular}{l|llll}
\toprule 
dataset & Plane & dihedral & trihedral & Plane+dihedral\\
&&&&+trihedral \\
\midrule 
ABC     &0.0063     & 0.0085    & 0.0078  & \textbf{0.0051}              \\
ScanNet & 0.0039    & 0.0057    & 0.0046    & \textbf{0.0033}  \\
\bottomrule
\end{tabular}
\caption{Ablation study of anglehedral surface.}
\label{tab:ablation}
\end{table}
However, how much the 3 geometry priors contributes separately is still a problem. Thus, we provide the visualization result to obtain a deeper sight in Fig.~\ref{fig:ablation_abc} and Fig.~\ref{fig:ablation_scan}. First, we focus on the sharp feature problem in Fig.~\ref{fig:ablation_abc}, we can see that dihedral surface and trihedral surface cannot work well independently and will lead to some holes on the flat surface due to its flexibility. Plane itself can generate rather good result on the flat surface. However, in sharp turning surface and boundary surface, it produces unsatisfactory result. The combination of the 3 geometry prior can produce best result. Next, we focus on the boundary problem in Fig.~\ref{fig:ablation_scan}, we can see that the method with only plane will expand outward near the boundary for the reason that the geometry prior plane cannot fit the the half plane near the boundary. From the ablation study, we can conclude that anglehedral surface have priority on representation of turning surface and boundary surface.

In addition, we provide the ablation result for 3 losses in Tab.~\ref{tab:ablation_loss} to clarify the affection for each loss. We choose ABC as an example and the number of points in a point cloud is $3000$. We can see that $L_{CD}$ is the most important one, and the other 2 losses take the auxiliary affection.  With the 3 losses, we achieve the best result. 
\begin{table}[]
\begin{tabular}{l|lll}
\toprule
Loss   & $L_{CD}$         & $L_{local}$         & $L_{UDF}$                   \\

$CD_1$ &     0.0061     &    0.0336     &       0.0475    \\
\midrule
Loss   & $L_{CD}+L_{UDF}$ & $L_{UDF}+L_{local}$ &                    \\
$CD_1$ &        0.0058          &  0.0349   &             \\
\midrule
Loss   & $L_{CD}+L_{local}$ & $L_{CD}+L{local}+L_{UDF}$ &\\
$CD_1$ &            0.0056      &   0.0051 &              \\
\bottomrule
\end{tabular}
\caption{Evaluation of the effectiveness of 3 losses on ABC.}
\label{tab:ablation_loss}
\end{table}

\subsection{Robust Analysis}
In this part, we provide the analysis of the important hyperparameters. First, we analyze the robustness of $k_1$ in Tab.~\ref{tab:robust_k1}. We can see that when $k_1$ is 36, the performance reaches the peak. Meanwhile, when $k_1$ varies in $[12, 24, 36, 48]$, the performance is relatively robust. 
\begin{table}[]
\centering
\begin{tabular}{l|llll}
\toprule 
$k_1$  & 12 & 24 & 36   & 48   \\
\midrule 
$CD_1$ & 0.0056 & 0.0053 & \textbf{0.0051} & 0.0.0052 \\
\bottomrule
\end{tabular}
\caption{The robust analysis of $k_1$ on ABC 001.}
\label{tab:robust_k1}
\end{table}

Next, we provide the analysis of the parameter in merging the local surface, $\theta$ in Tab.~\ref{tab:robust_theta}. We can see that when $\theta$ varies in $[50, 100, 200, 400]$, the performance varies a little. When $\theta$ is 100, the performance reaches the peak.
\begin{table}[]
\centering
\begin{tabular}{l|llll}
\toprule 
$\theta$  & 50 & 100 & 200   & 400   \\
\midrule 
$CD_1$ & 0.0053 & 0.0051 & 0.0052 & 0.0054 \\
\bottomrule
\end{tabular}
\caption{The robust analysis of $\theta$ on ABC 001}
\label{tab:robust_theta}
\end{table}

\subsection{Efficiency Analysis}
We only add anglehedral surface comparing with SuperUDF\cite{superudf}. Obviously, the representation is more complex than plane. In this part, we provide the time comparison between plane representation and anglehedral surface representation during training and testing part in Tab~\ref{tab:efficiency}. We can see that although anglehedral surface is more complex, the time cost during training and testing increase a little. This is for the reason that the complexity increased by the anglehedral surface is too low in modern computer. 
\begin{table}[]
\centering
\begin{tabular}{l|llll}
\toprule 
Method              & NPull & CAP & SuperUDF & Ours \\
\midrule 
field &&&&\\
prediction & 29min22s     & 21min56s   & 9s       & 10s   \\
\bottomrule
\end{tabular}
\caption{Time Efficiency comparison between ours and other method of (signed or unsigned) distance prediction.}
\label{tab:efficiency}
\end{table}

In Tab.~\ref{tab:param}, we can see the FLOP and the number of parameters comparison. We can see that although FLOP and the number of parameters of anglehedral surface is more than plane, they are both very small. Here, we only calculate the memory and FLOP of geometry priors prediction part and cut down the calculation cost of backbone network.


\begin{table}[]
\centering
\begin{tabular}{l|ll}
\toprule 
Method     & SuperUDF& plane+dihedral\\
&&+trihedral \\
\midrule 
\#parameter(MB) & 0.126  & 0.230                 \\
\#FLOP(G)      &8.097  & 10.271            \\
\bottomrule
\end{tabular}
\caption{The number of parameter and FLOP of the anglehedral surface. The number of point cloud is 3000 and the number of query point is 6000}
\label{tab:param}
\end{table}

\subsection{More Visualization Result}
\label{app:vis}
\begin{figure*} 
\centering
  \includegraphics[width=1.8\columnwidth]{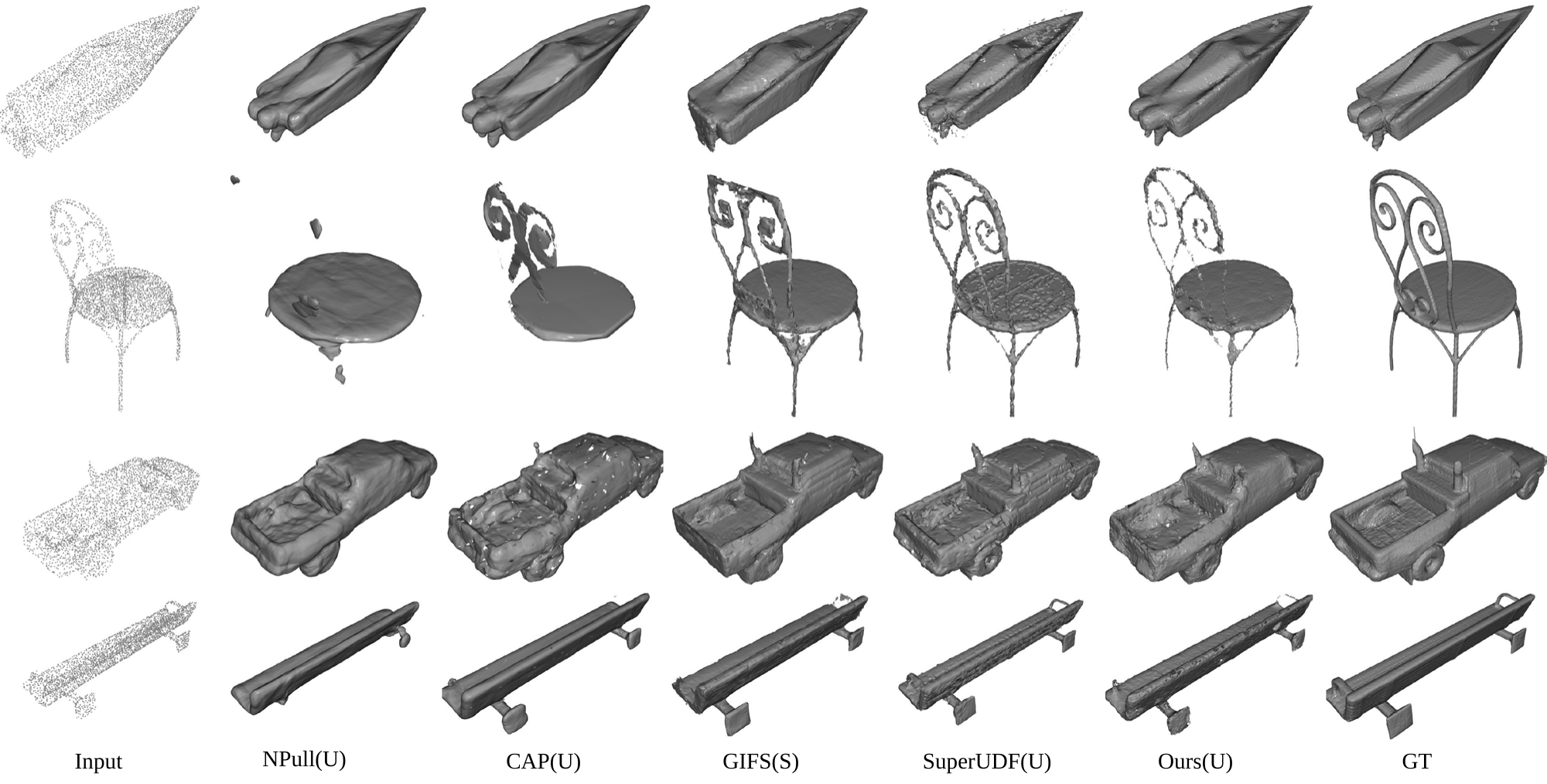}
  \caption{Visualized results of our method and state-of-the-art alternatives on ShapeNet. (S) mean supervised method and (U) means unsupervised method. Methods are NPull, CAP, GIFS, SuperUDF and Ours. The number of input point cloud is 3000.}
  \label{fig:shapenet_result2}
\end{figure*}

\begin{figure*} 
\centering
  \includegraphics[width=1.8\columnwidth]{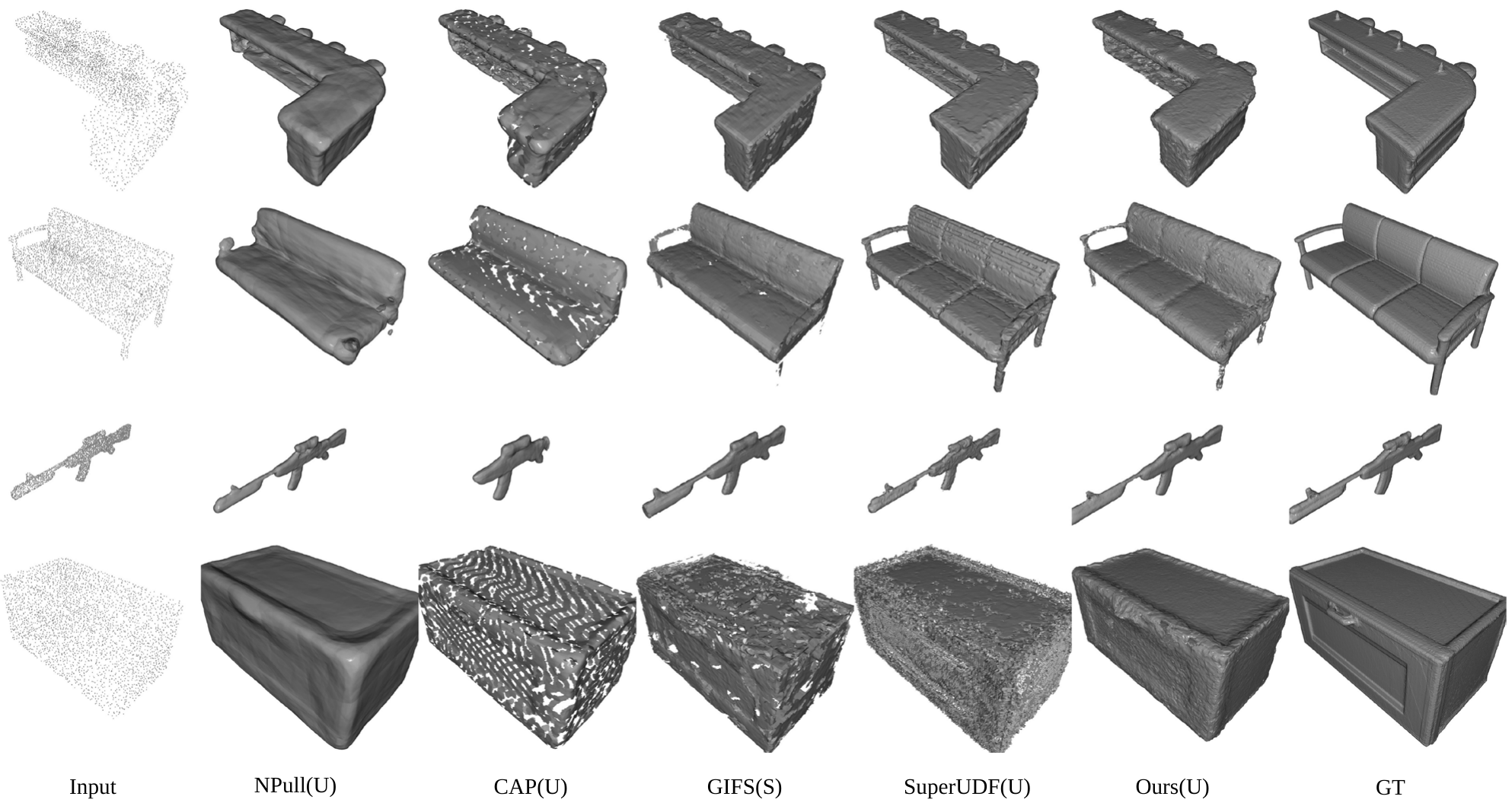}
  \caption{Visualized results of our method and state-of-the-art alternatives on ShapeNet. (S) mean supervised method and (U) means unsupervised method. Methods are NPull, CAP, GIFS, SuperUDF and Ours. The number of input point cloud is 3000.}
  \label{fig:shapenet_result3}
\end{figure*}

More visualization result on ShapeNet dataset can be found in Fig.~\ref{fig:shapenet_result2} and Fig.~\ref{fig:shapenet_result3}.

\begin{figure*} 
\centering
  \includegraphics[width=1.8\columnwidth]{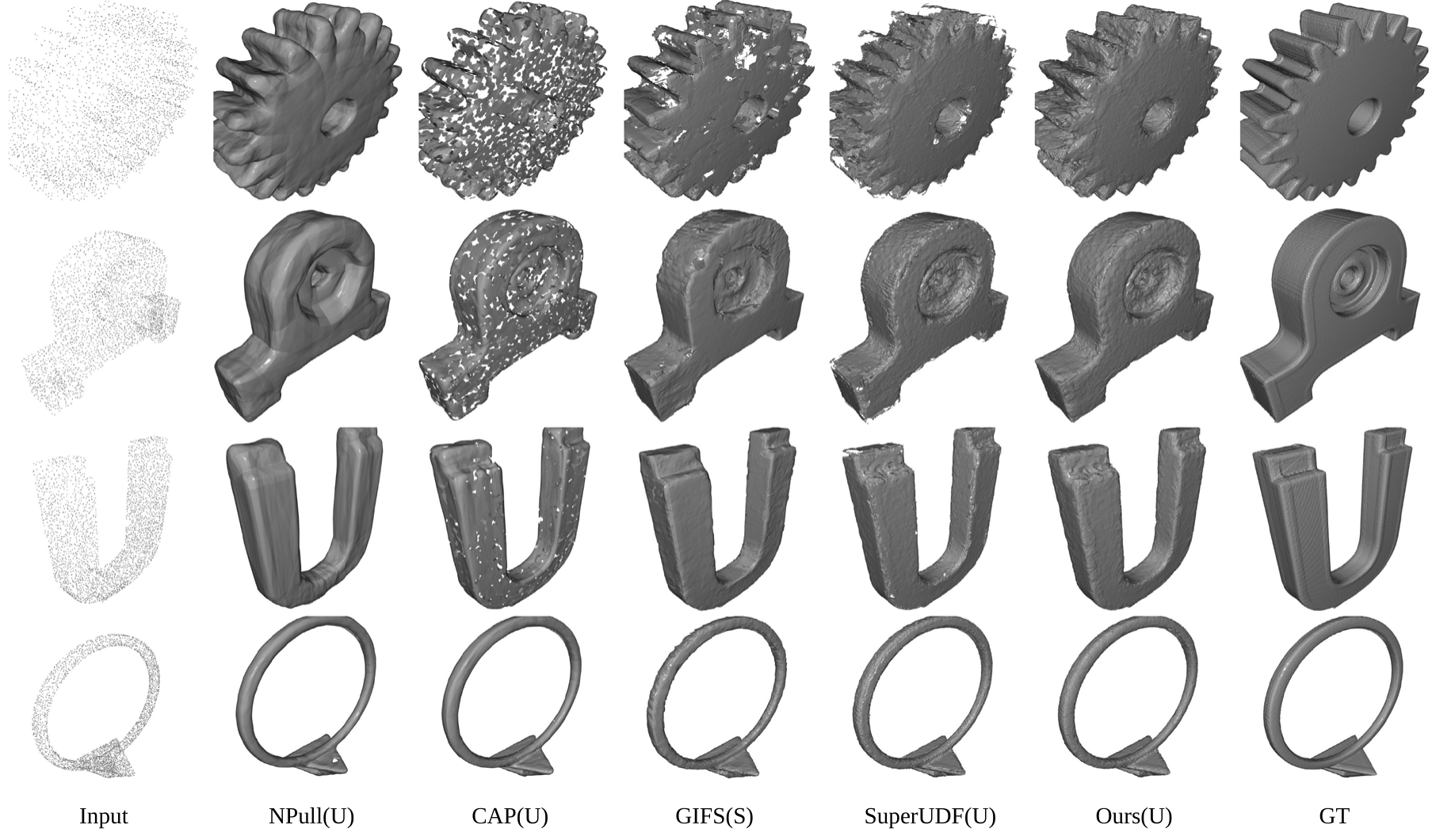}
  \caption{Visualized results of our method and state-of-the-art alternatives on ABC 001. (S) mean supervised method and (U) means unsupervised method. Methods are NPull, CAP, GIFS, SuperUDF and Ours. The number of input point cloud is 3000.}
  \label{fig:abc_result1}
\end{figure*}

\begin{figure*} 
\centering
  \includegraphics[width=1.8\columnwidth]{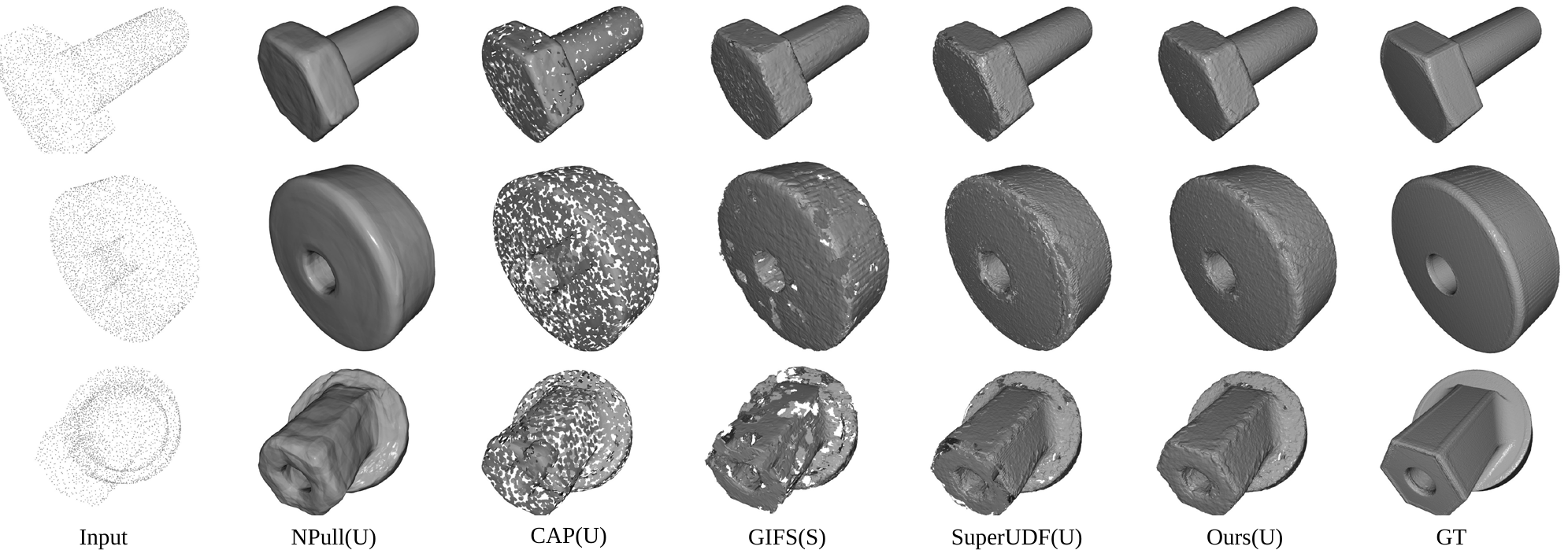}
  \caption{Visualized results of our method and state-of-the-art alternatives on ABC 001. (S) mean supervised method and (U) means unsupervised method. Methods are NPull, CAP, GIFS, SuperUDF and Ours. The number of input point cloud is 3000.}
  \label{fig:abc_result3}
\end{figure*}

More visualization result on ABC dataset can be found in Fig. 
\ref{fig:abc_result1} and Fig. \ref{fig:abc_result3}.

\begin{figure*} 
\centering
  \includegraphics[width=1.8\columnwidth]{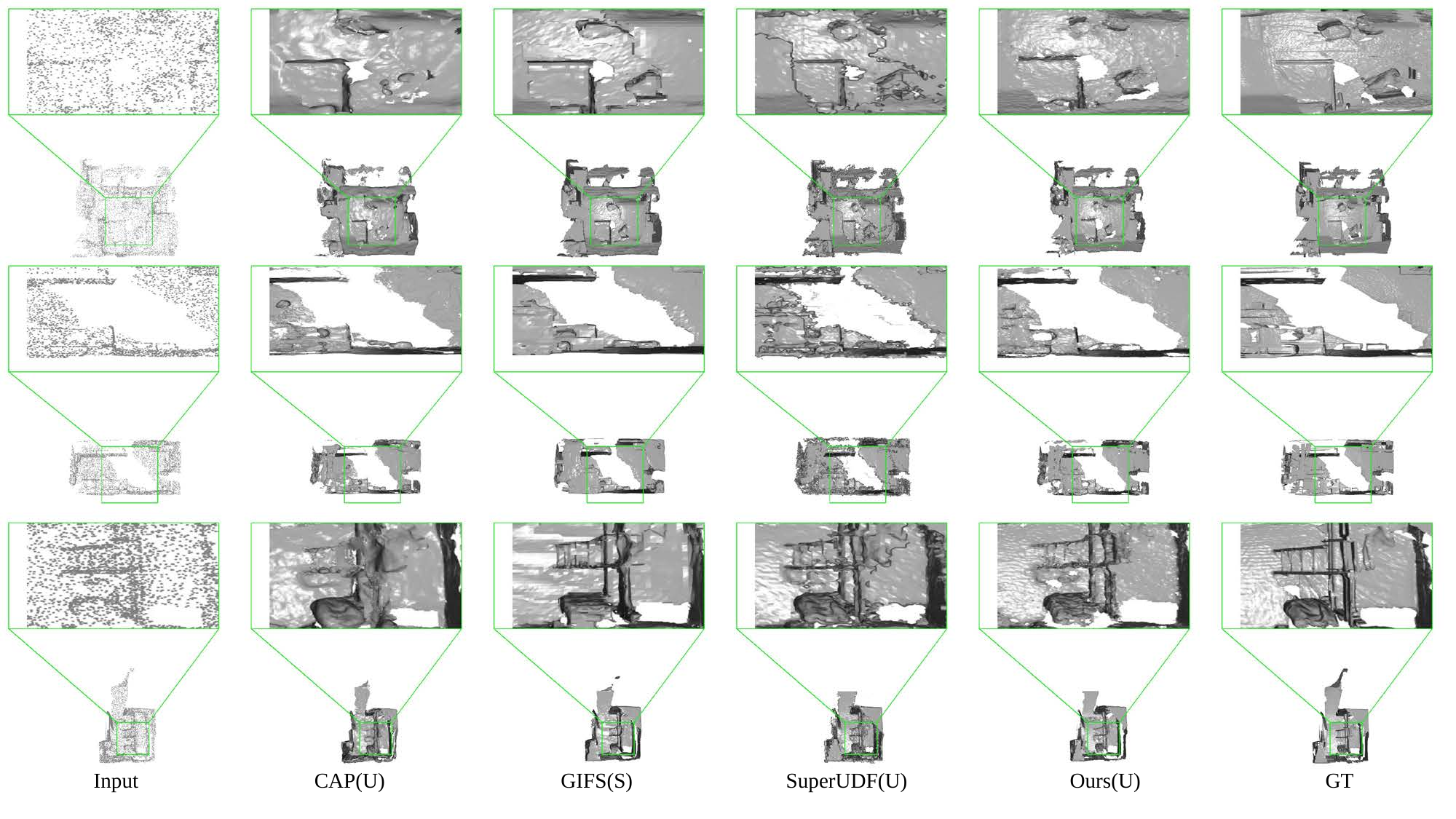}
  \caption{Visualized results of our method and state-of-the-art alternatives on ShapeNet. (S) mean supervised method and (U) means unsupervised method. Methods are CAP, GIFS, SuperUDF and Ours. The number of input point cloud is 10000.}
  \label{fig:scan_result2}
\end{figure*}

More visualization result on ScanNet dataset can be found in Fig. \ref{fig:scan_result2}.

\section{Conclusion}
We introduce the concept of an anglehedral surface to characterize local surfaces, encompassing both dihedral and trihedral surfaces. This representation effectively preserves the sharp features of turning surfaces, cube corners, and boundary surfaces. The anglehedral surface relies solely on normals, eliminating the need for a local coordinate system and thereby circumventing coordinate consistency issues. However, challenges persist in the representation. For instance, it struggles with noisy input due to its local perspective, making it difficult to strike a balance between sharp features and noise interference. In addition, it is hard for our method to handle the point cloud with its density varying a lot. Future work will focus on the problem. 

\section*{Availability of data and materials}
The ABC dataset is available from
https://archive.nyu.edu/jspui/handle/2451/43778. ShapeNet dataset is available from  http://www.shapenet.org. ScanNet dataset is available from http://www.scan-net.org/.

\section*{Competing interests}
The authors have no competing interests to declare that
are relevant to the content of this article. The author
Kai Xu is the Area Executive Editor of this journal.

\section*{Funding}
This work is supported in part by the National Natural Science Foundation of China
(62325211, 62132021).

\section*{Authors' contributions}
Hui Tian: Methodology, Writing Draft, Visualization,
Results Analysis;  
Kai Xu: Writing Revision, Supervision.

\section*{Acknowledgements}
We thanks the GIFS\cite{gifs}, it is an excellent work and inspires us a lot.

\section*{Author information}

\begin{biography}[TH]{Hui Tian}  received his B.E. and
M.E. degrees in computer science and
technology from NUDT in 2017 and
2019, respectively. He is currently
pursuing a Ph.D. degree. His research
interests include
surface reconstruction and 3D representation learning.
\end{biography}
\vspace*{0.3em}

\begin{biography}[Kai_Xu]{Kai XU}  Kai Xu is a professor in the School of
 Computing, NUDT, where he received
 his Ph.D. degree in 2011. He serves on
 the editorial board of ACM Transactions
 on Graphics, Computer Graphics Forum,
 Computers \& Graphics, etc.
\end{biography}

\bibliographystyle{CVMbib}
\bibliography{bibfile}


\end{document}